%% file: main.tex
\documentclass[10pt,letterpaper]{article}

\usepackage[letterpaper,margin=1in,columnsep=0.28in]{geometry}
\usepackage[utf8]{inputenc}
\usepackage[T1]{fontenc}
\usepackage{newtxtext,newtxmath}
\usepackage{amsmath,amsfonts}
\usepackage{booktabs}
\usepackage{graphicx}
\usepackage{multirow}
\usepackage{array}
\usepackage{nicefrac}
\usepackage{microtype}
\usepackage{xcolor}
\usepackage{xspace}
\usepackage{tikz}
\usepackage{pgfplots}
\usepackage{algorithm}
\usepackage{algorithmic}
\usepackage{float}
\usepackage{enumitem}
\usepackage[numbers,sort&compress]{natbib}
\usepackage[font=small,labelfont=bf]{caption}
\usepackage[breaklinks=true,colorlinks=true,citecolor=blue,linkcolor=blue,urlcolor=blue]{hyperref}
\pgfplotsset{compat=1.17}

\setlist[enumerate]{leftmargin=1.35em,itemsep=0.2em,topsep=0.25em}
\newcommand{\sys}{\emph{InViStream}\xspace}

\title{Cloak of Invisibility: Real-Time Privacy-Preserving Volumetric Video Streaming}
\input{authors.tex}
\date{}

\hypersetup{
  pdftitle={Cloak of Invisibility: Real-Time Privacy-Preserving Volumetric Video Streaming},
  pdfauthor={Hossein Khalili, Philip Do, Alexander Vilesov, Achuta Kadambi, Kittipat Apicharttrisorn, Nader Sehatbakhsh},
  pdfsubject={Privacy-preserving volumetric video streaming},
  pdfkeywords={volumetric video, privacy, RGB-D, multi-view streaming, edge computing}
}

\begin{document}
\makeatletter
\twocolumn[
\begin{@twocolumnfalse}
\maketitle
\begin{abstract}
Volumetric video streaming turns privacy into a 3D, multi-view problem. Unlike ordinary video, where sensitive content can often be redacted frame by frame, RGB-D volumetric pipelines capture people, rooms, and personal objects from multiple cameras and then fuse them into a shared 3D representation. A private object missed in one view, or only partially removed before fusion, can therefore reappear in the reconstructed scene. This creates a new privacy challenge for 3D telepresence, education, entertainment, and immersive applications: private content should be removed before raw visual and geometric data leave the camera side, while the public part of the scene should remain useful for real-time reconstruction. Existing volumetric streaming systems mainly optimize reconstruction, data movement, and latency, while privacy-preserving vision methods are typically designed for single-camera, single-frame images and do not directly address calibrated multi-view RGB-D fusion. We present InViStream, a real-time ``privacy-from-source'' system explicitly designed for this setting. InViStream addresses three challenges that arise in volumetric capture: private objects may appear differently across views, RGB masking alone can leave geometric privacy leakage in depth, and public/private instances of the same class must be separated consistently before cloud-side fusion. To address these challenges, InViStream combines standard object detection with depth-aware masking, propagates public/private decisions across calibrated views, and fuses only sanitized point clouds. We evaluate InViStream on diverse synthetic and real RGB-D scenes, including offices, conference rooms, living rooms, and settings with multiple public and private people and objects. Across these scenarios, InViStream achieves an average synthetic Dice/Recall of 0.799/0.891 and a real Dice/Recall of 0.792/0.908, keeps synthetic SSIM above 0.98, and supports real-time streaming above 30 FPS.
\end{abstract}
\vspace{0.8em}
\end{@twocolumnfalse}
]
\makeatother

\section{Introduction}
Volumetric video streaming (VVS) represents scenes as time-varying 3D content, usually by converting synchronized RGB-D views into point clouds or meshes. 

This enables six degrees of freedom telepresence, remote collaboration, interactive education, and mixed-reality experiences. Recent VVS systems have improved throughput, bandwidth, reconstruction quality, and headset rendering \citep{orts2016holoportation,wu2024theia,zhang2024habitus,zhang2022m5,zhang2021efficient,guan2023metastream,lee2023farfetchfusion,liu2022vues,han2020vivo,chen2024immerscope,cheng2024magicstream,liu2024muv2}. As these systems move from controlled demos to offices, homes, and shared spaces, they also inherit a more difficult privacy problem than conventional 2D video: wide-space capture can include background people, screens, documents, photographs, and personal objects from many viewpoints.

The privacy failure happens early in the pipeline. In many architectures, raw RGB-D frames are sent to a server for reconstruction, compression, or streaming. Once the cloud or remote user receives those frames, later blurring or access control cannot undo the initial exposure. A practical VVS privacy system therefore needs \emph{privacy from source}: private content should be filtered before it leaves the camera side, while the public parts of the scene remain useful enough for volumetric reconstruction.

Existing approaches do not directly solve this setting. Prior VVS systems \textit{only} optimize delivery performance and immersion, and do not even consider the privacy problem. Existing privacy-preserving image analysis has explored local transformations, pixelation, obfuscation, and secure inference \citep{wu2021pecam,lu2022preva,yu2018pinto,singh2021disco,padilla2015visual,mireshghallah2020shredder,van2022client,jiang2022primask,kim2017viewmap,ilia2015face,zhu2023campro,shan2020fawkes}. However, these methods \textit{only} operate on one camera or one 2D frame. They are not designed to handle same-class ambiguity (one person is public, another is private), depth-dependent masking, or cross-view consistency during 3D fusion. As a result, they achieve low recall and/or suffer from high latency.

\begin{figure}[t]
  \centering
  \includegraphics[width=0.9\linewidth]{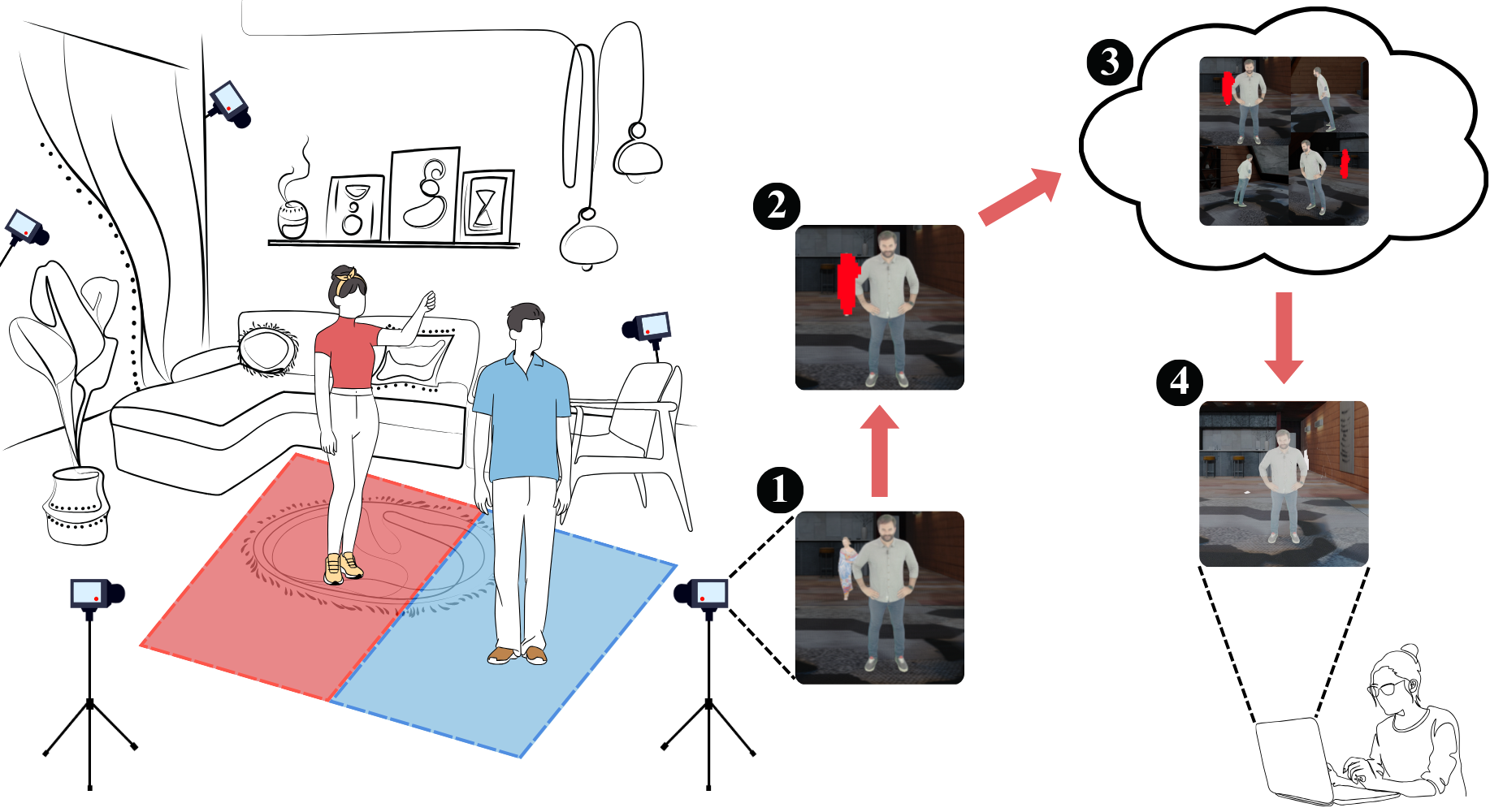}
  \caption{\textbf{Privacy-from-source volumetric streaming.} RGB-D cameras capture a wide space, edge devices mask user-defined private objects before transmission, the cloud fuses only sanitized views, and the remote user receives a point cloud in which private objects are removed.}
  \label{fig:intro}
\end{figure}

This paper introduces \sys, a real-time privacy-preserving VVS pipeline. The key idea is to combine three pieces of information that are individually insufficient but jointly effective: object identity from a lightweight detector, object geometry from depth, and multi-view consistency from calibrated camera transforms. \sys keeps the detector off the critical path by running it only on selected frames, then reuses depth profiles to mask intervening frames. It also treats privacy as an instance-level policy, not as a class-level policy: the user can keep the foreground participant public while removing background people or objects of the same semantic class. An overview of our approach is shown in Figure~\ref{fig:intro}.

Our contributions are:
\begin{enumerate}[leftmargin=1.3em,itemsep=0.2em,topsep=0.2em]
  \item We formulate source-side privacy for calibrated RGB-D volumetric video under an honest-but-curious cloud model, including the public/private instance policy needed for same-class objects.
  \item We design \sys, a depth-aware multi-view masking pipeline that combines off-the-shelf detection, depth-conditioned mask construction, reference-view public/private synchronization, and private point removal before cloud-side fusion.
  \item We evaluate \sys on synthetic and real RGB-D scenes with multiple public/private objects, measuring privacy, utility, and real-time performance. \sys achieves average synthetic Dice/Recall of 0.799/0.891, average real Dice/Recall of 0.792/0.908, and large reductions in private-person detection after masking.
\end{enumerate}

\section{Problem Setup and Threat Model}

\paragraph{Application focus.} \sys is intended for wide-space RGB-D capture, where intentional participants may share the scene with incidental private content. This distinguishes the task from object-level volumetric streaming and ordinary 2D redaction: privacy failures can arise from missed camera views, remaining depth geometry, or inconsistent public/private decisions before fusion. We therefore organize the paper around the requirements of a deployed application: private-object removal, public-scene preservation, same-class public/private ambiguity, and interactive runtime.

\paragraph{Volumetric setting.} We consider a room-scale VVS system with $V$ calibrated RGB-D cameras. At time $t$, camera $v$ captures an RGB image $I_{v,t}$ and an aligned depth map $D_{v,t}$. A local edge device near each camera can run lightweight processing, but expensive 3D fusion is delegated to a cloud server. The desired output is a merged point cloud $\mathcal{P}_t$ that preserves public content and removes private instances.

\paragraph{Privacy policy.} A user defines a set of sensitive classes $\mathcal{C}$, such as people, screens, phones, or picture frames. Privacy is instance-level: some objects in a sensitive class may be public while other same-class objects are private. \sys uses a reference view and a user-defined working area to identify public instances, such as the participant in the foreground of a telepresence session. All sensitive instances outside that policy are private. This policy is intentionally simple and auditable; it can be replaced by another user interface without changing the core pipeline.

\paragraph{Adversary and goal.} We assume an honest-but-curious cloud and remote receiver. They execute the protocol but may inspect any RGB, depth, mask-color, or point-cloud data they receive. The edge device and camera are trusted to apply masking before transmission. The privacy goal is to prevent private object appearance and geometry from reaching the cloud or the receiver, to the extent supported by detector and depth quality. The utility goal is to preserve public geometry and visual quality. We do not claim cryptographic privacy and do not address compromised cameras, malicious edge devices, audio leakage, or adversarial physical attacks on the detector. These scope boundaries are important because \sys targets practical low-latency source-side filtering rather than secure computation.

\begin{table}[t]
\centering
\caption{Threat-model summary used throughout the paper.}
\label{tab:threat}
\small
\begin{tabular}{p{0.23\linewidth}p{0.66\linewidth}}
\toprule
Component & Assumption and role \\
\midrule
Trusted source & Camera-side device applies detection, depth masking, and privacy policy before transmission. \\
Untrusted cloud & receives only sanitized RGB-D frames or sanitized point clouds; it may inspect data, but follows the fusion protocol. \\
Private assets & RGB appearance, depth geometry, and 3D points belonging to user-defined private instances. \\
Out of scope & Compromised sensors, malicious edge devices, audio/metadata leakage, adversarial examples, and legal policy enforcement. \\
\bottomrule
\end{tabular}
\end{table}

\section{InViStream Design}

\begin{figure*}[t]
  \centering
  \includegraphics[width=0.98\linewidth]{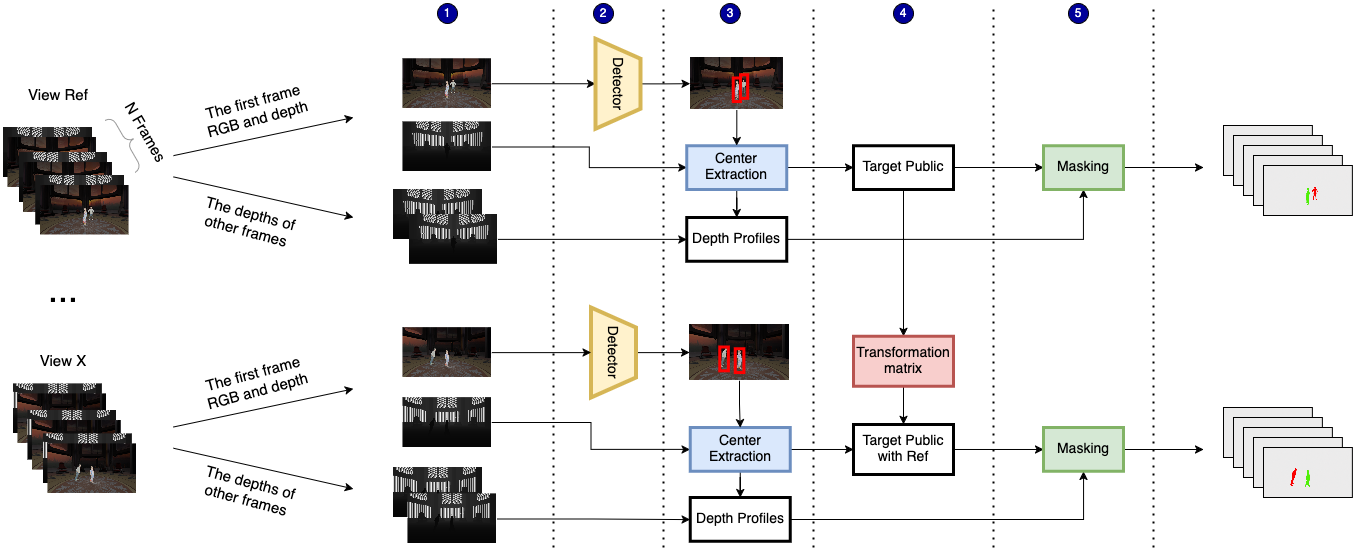}
  \caption{\textbf{Depth-aware multi-view masking.} The detector runs on the first frame of each chunk. \sys extracts depth profiles, identifies public instances in a reference view, transfers public centers to other views with calibrated transforms, and masks all non-public sensitive instances using depth and bounding-box constraints.}
  \label{fig:pipeline}
\end{figure*}

Figure~\ref{fig:pipeline} summarizes the pipeline. \sys processes each camera stream locally, transmits sanitized RGB-D data, and uses the cloud only for point-cloud conversion, private point deletion, alignment, and streaming.

\paragraph{Object detection on chunk starts.} Running instance segmentation on every camera frame is expensive for edge devices. \sys instead uses an off-the-shelf detector on the first frame of a chunk of length $N$. In our implementation, we use Faster R-CNN with ResNet-50-FPN or MobileNetV3 backbones \citep{ren2015faster,he2016deep,lin2017feature,howard2019searching}, and we also evaluate Tiny YOLOv2-style lightweight detection \citep{Redmon2017YOLO9000}. The detectors are pretrained on COCO \citep{lin2014microsoft} and are not fine-tuned on our synthetic or real scenes. For a detection frame, the detector returns boxes $\mathcal{O}_{v,t}=\{(b_i,s_i,l_i)\}_{i=1}^{n_t}$. Boxes with $s_i \ge \theta$ and $l_i \in \mathcal{C}$ become sensitive candidates.

\paragraph{Depth-conditioned masks.} A detector box alone is not a privacy mask: masking the entire box removes too much public content, while generic segmentation may not identify which same-class instance is private. For each sensitive box $b_i=(x_i,y_i,w_i,h_i)$, \sys computes the box center
\begin{equation}
  c_x^i=x_i+\frac{w_i}{2},\qquad c_y^i=y_i+\frac{h_i}{2}.
\end{equation}
It extracts a $w\times w$ depth window $V_i$ around $(c_x^i,c_y^i)$ and computes a depth profile
\begin{equation}
\begin{aligned}
  \mu_i&=\frac{1}{|V_i|}\sum_{d\in V_i}d, \\
  \sigma_i&=\sqrt{\frac{1}{|V_i|}\sum_{d\in V_i}(d-\mu_i)^2},\qquad
  \tau_i=\alpha\sigma_i .
\end{aligned}
\end{equation}
Pixels inside the candidate box are masked if their depth is consistent with the object profile:
\begin{equation}
  M_i(u,v)=\mathbf{1}\{(u,v)\in b_i\}\cdot\mathbf{1}\{|D_{v,t}(u,v)-\mu_i|\le \tau_i\}.
\end{equation}
The bounding-box term prevents depth-similar background regions from being removed, while the depth term avoids coarse box-level overmasking.

\paragraph{Public/private synchronization across views.} The reference view defines the public instances for each sensitive class, usually by selecting objects inside a user-specified working area. For each public object center $(r_x,r_y,r_z)$ in reference coordinates, \sys projects the center to a target view $w$ using the calibrated transform $\mathbf{T}_{r\rightarrow w}$:
\begin{equation}
\begin{pmatrix} T_x&T_y&T_z&1 \end{pmatrix}^{\top}
=\mathbf{T}_{r\rightarrow w}
\begin{pmatrix} r_x&r_y&r_z&1 \end{pmatrix}^{\top}.
\end{equation}
In the target view, the detected same-class box whose center is closest to the projected point is labeled public; all other same-class boxes are labeled private. This rule is conservative for privacy: an object not matched to a declared public instance is masked. It also handles cases where a private object is absent in the reference view but visible elsewhere, because that object will not be matched to any transferred public center.

\paragraph{Point removal and cloud fusion.} After local masking, sanitized RGB-D frames are sent to the server. The server converts each view to a point cloud, removes points whose color corresponds to the private mask, transforms each point cloud to the global coordinate frame, and merges the remaining points. In our implementation, private points are assigned non-finite coordinates and removed using Open3D's optimized non-finite-point filtering \citep{open3d}. The rest of the volumetric streaming pipeline remains compatible with existing VVS systems: \sys changes what data is allowed to enter the cloud, not the downstream renderer.

\paragraph{Temporal optimization.} For frames between detector calls, \sys reuses the stored depth profiles and bounding boxes. This reduces detection overhead but creates a privacy/latency trade-off: if a private object moves too far during a chunk, recall can degrade. Section~\ref{sec:robustness} and Appendix~\ref{app:robustness} quantify this effect.

\section{Evaluation}

We evaluate whether \sys preserves privacy, keeps public content useful, and runs fast enough for interactive streaming.

\paragraph{Datasets.} The synthetic dataset uses three open-source 3D scenes modified with Open3D \citep{open3d,sketch}. We insert public and private human models from V-Sense and 8i datasets \citep{zerman2019subjective,krivokuca2018voxelized}, render eight RGB-D views per scene, and generate instance-level ground truth from the known 3D models. The resulting benchmark contains more than 400 scene configurations with varied backgrounds, viewpoints, object distances, overlap, and occlusion. The real dataset uses an Intel RealSense D435 depth camera across five representative environments: conference room, open-plan office, floor corridor, outdoor patio, and living room. Each environment is captured from eight calibrated viewpoints under four activities. The real dataset contains more than 150 scene instances and includes six adult participants. 
All participants provided informed consent for the research use of their RGB-D captures and 
anonymized paper figures. Additional human-subjects, data-handling, and release details are 
provided in Appendix~\ref{app:implementation}. To avoid time-synchronization artifacts during measurement, real multi-view captures are collected sequentially in static capture epochs; Section~\ref{sec:limitations} discusses this limitation. Table~\ref{tab:evaluation_coverage} summarizes the evaluation coverage that stresses the main failure modes for privacy-from-source volumetric streaming.

\begin{table}[t]
\centering
\caption{Evaluation coverage. The benchmark stresses multi-view geometry, same-class public/private ambiguity, and real RGB-D sensing artifacts.}
\label{tab:evaluation_coverage}
\small
\begin{tabular}{@{}p{0.20\linewidth}p{0.32\linewidth}p{0.34\linewidth}@{}}
\toprule
Subset & Coverage & Main challenge \\
\midrule
Synthetic rooms & 3 room meshes, 8 virtual views each, 400+ configurations & depth variation, occlusion, and public/private placement \\
Human models & 8 3D human models, up to 2 public and 2 private people & same-class instance ambiguity \\
Real scenes & 5 environments, 8 calibrated viewpoints, 150+ instances & clutter, lighting, mirrors, glass, and depth jitter \\
Activities & conversation, walking, sitting, and standing & motion between detector refreshes \\
\bottomrule
\end{tabular}
\end{table}

\paragraph{Metrics.} We report Dice, recall, SSIM, point distance, latency, FPS, and private-object detection rate (PODR). Dice measures private-mask overlap with ground truth. Recall measures how much of the private object is removed and is therefore the most direct privacy-side mask metric. SSIM and point distance measure public scene quality after private-object removal. PODR measures whether a detector still finds the private person after masking; lower is better.

\begin{figure}[t]
  \centering
  \includegraphics[width=0.98\linewidth]{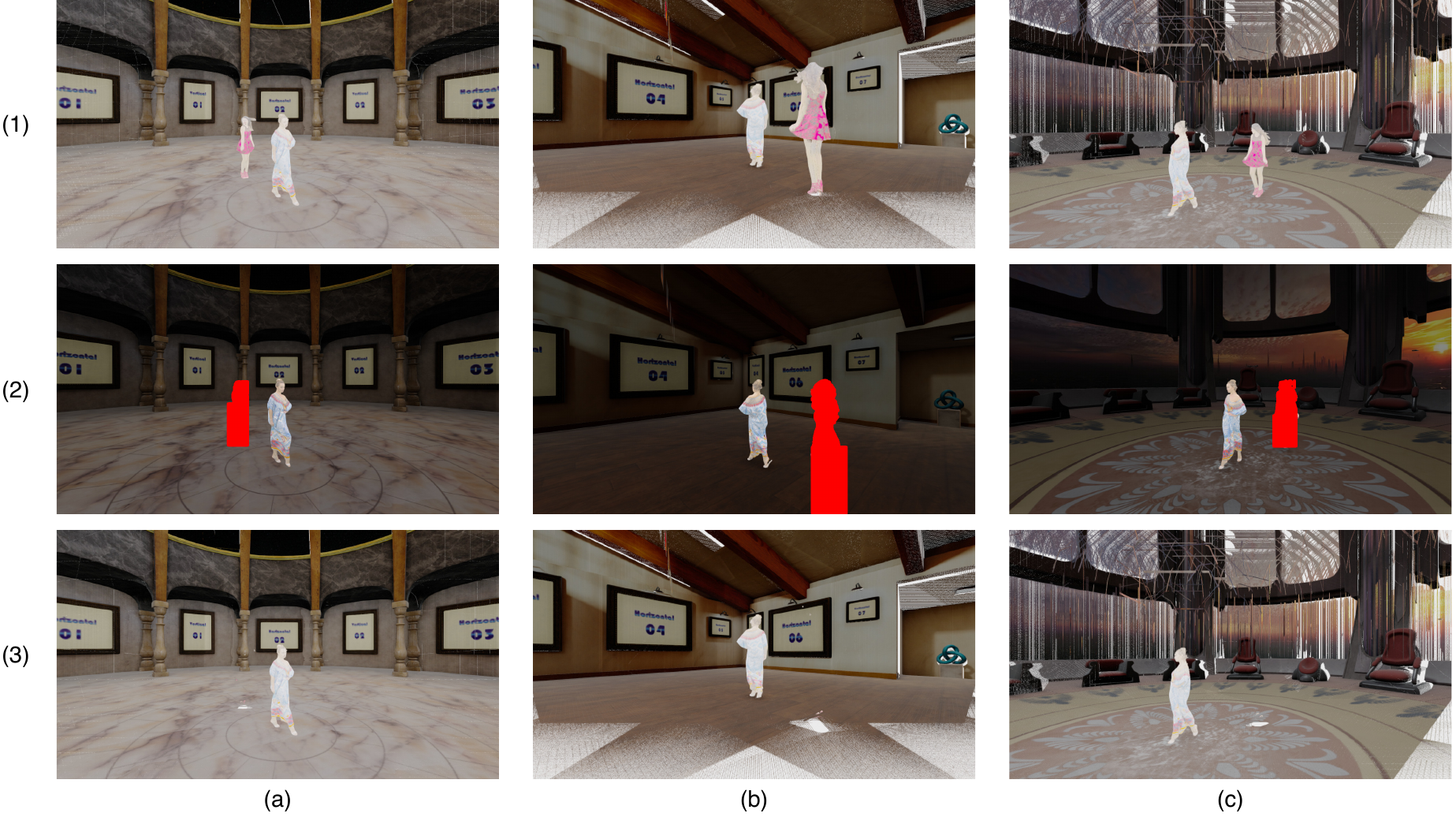}
  \caption{Qualitative synthetic results. Row 1 shows the original scene, row 2 shows edge-side private masks in red, and row 3 shows the final 3D scene after removing private points. \sys removes private people while preserving the public person and background content across different room geometries.}
  \label{fig:qual}
\end{figure}

\paragraph{Synthetic scenes.} Figure~\ref{fig:qual} shows representative outputs. Table~\ref{tab:main_results} reports quantitative results across the three synthetic scenes. Dice stays near 0.8, recall is at least 0.885, SSIM remains above 0.98, and point distance is close to zero. These results indicate that \sys removes private objects without substantially changing public 3D content.

\begin{table}[t]
\centering
\caption{Main accuracy and utility results. Synthetic scenes include SSIM and point distance because exact 3D ground truth is available. Real scenes report Dice/Recall over all frames and views.}
\label{tab:main_results}
\small
\resizebox{\linewidth}{!}{%
\begin{tabular}{lcccc}
\toprule
Scene & Dice $\uparrow$ & Recall $\uparrow$ & SSIM $\uparrow$ & Point dist. $\downarrow$ \\
\midrule
Gallery House & 0.807 & 0.903 & 0.987 & 1.09e-4 \\
Gallery Round & 0.802 & 0.885 & 0.991 & 1.25e-4 \\
Jedi Council & 0.788 & 0.885 & 0.988 & 3.82e-4 \\
\midrule
\textbf{Synthetic average} & 0.799 & 0.891 & 0.989 & 2.05e-4 \\
\midrule
Conference room & 0.790 & 0.931 & n/a & n/a \\
Living room & 0.849 & 0.839 & n/a & n/a \\
Patio & 0.791 & 0.882 & n/a & n/a \\
Office & 0.736 & 0.981 & n/a & n/a \\
Floor corridor & 0.793 & 0.909 & n/a & n/a \\
\midrule
\textbf{Real average} & 0.792 & 0.908 & n/a & n/a \\
\bottomrule
\end{tabular}
}%
\end{table}

\begin{figure}[t]
  \centering
  \includegraphics[width=0.88\linewidth]{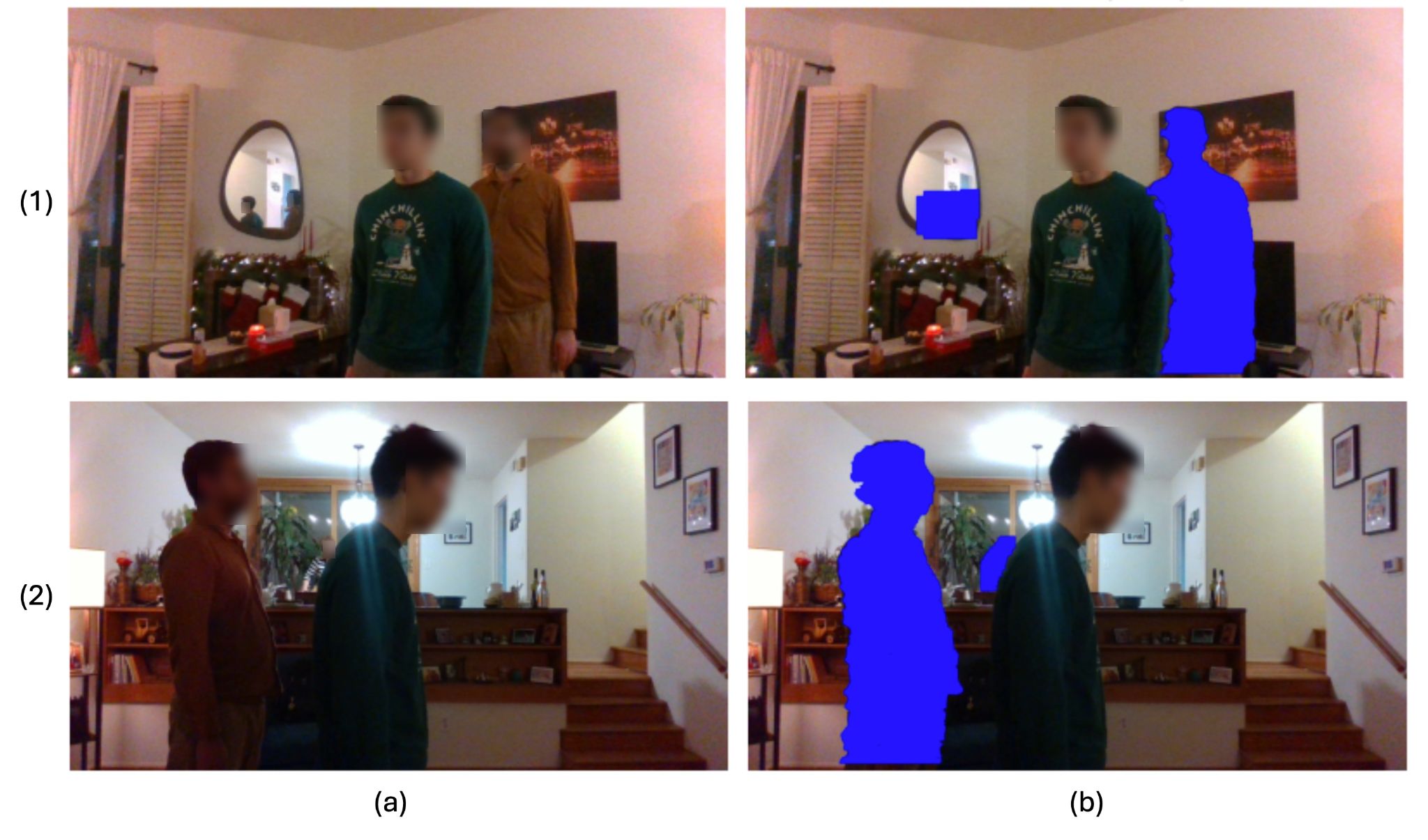}
  \caption{\textbf{Real-world hard cases.} Each row shows the original image, ground-truth mask, and predicted mask. In the first row, the mask covers both a person and the mirror reflection; in the second row, the mask covers a distant background person.}
  \label{fig:real_hard_main}
\end{figure}

\paragraph{Real scenes.} Real RGB-D data introduces depth jitter, imperfect calibration, lighting variation, mirrors, glass, and clutter. Even under these conditions, the real-scene average is 0.792 Dice and 0.908 Recall (Table~\ref{tab:main_results}). The office scene has the highest recall and lower Dice, indicating a privacy-favoring overmasking behavior; the living room has the highest Dice but lower recall because depth reflections and clutter make local depth profiles harder to estimate. This is the expected privacy/utility trade-off for a source-side system. Figure~\ref{fig:real_hard_main} shows two real hard cases from the original benchmark: a mirror reflection and a distant background person. These cases are useful because they stress two different parts of the pipeline: depth instability near reflective surfaces and small private objects at long range.

\begin{figure}[t]
  \centering
  \includegraphics[width=0.9\linewidth]{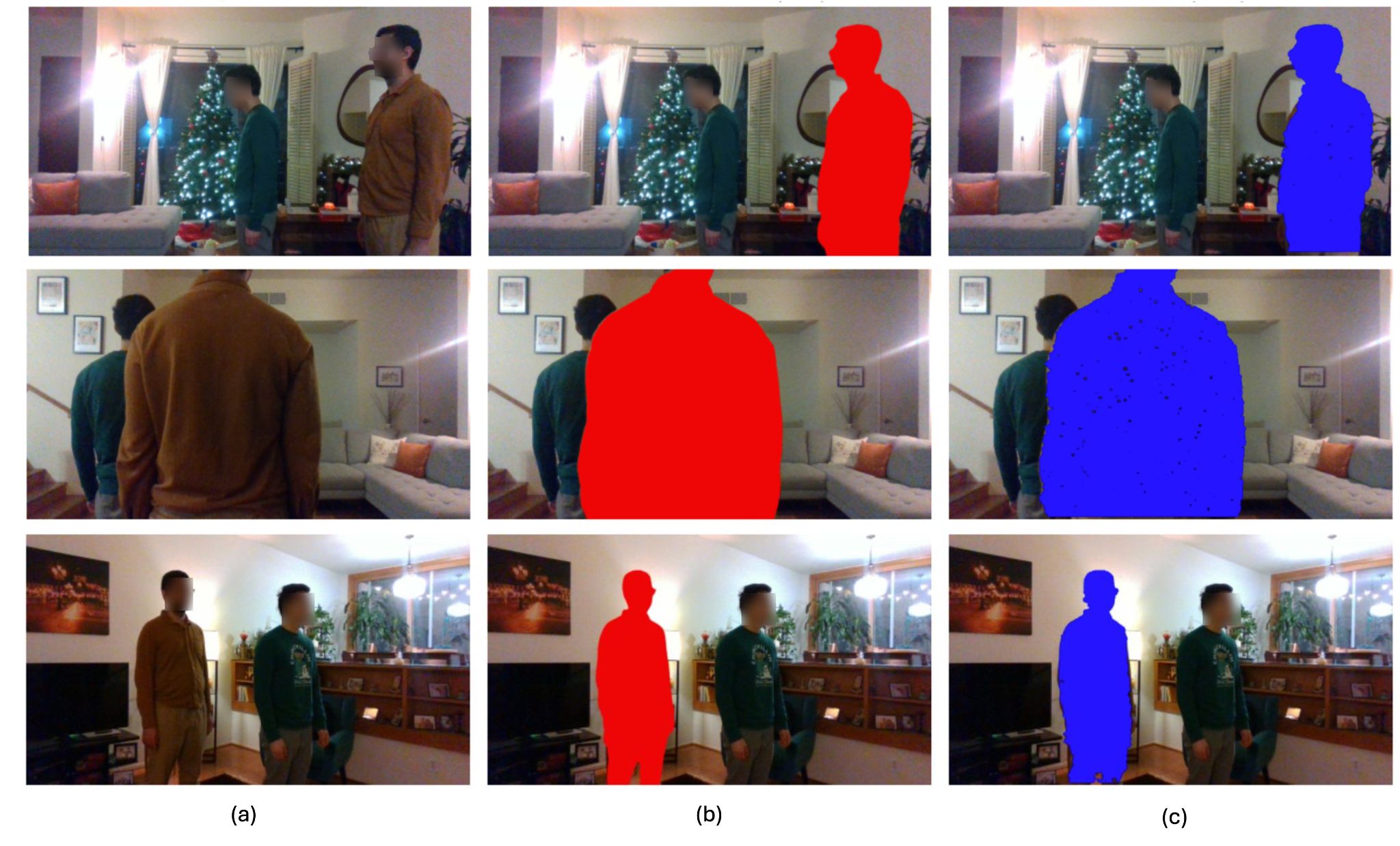}
  \caption{Additional real-scene qualitative example from the existing benchmark. The columns show the original RGB image, the ground-truth private-object mask, and the predicted mask.}
  \label{fig:real_qual_main}
\end{figure}

\paragraph{Mixed public/private crowds.} In scenes with three or more people, \sys remains robust but becomes more sensitive to calibration and matching errors. Table~\ref{tab:mixed_people_main} summarizes the crowd stress test from the synthetic benchmark. This is the most important instance-level privacy test because all people share the same detector label, so a class-level masker would either remove everyone or leak a private person. Recall remains high in the privacy-favoring cases, while Dice drops when transformed public centers are close to private boxes and the system conservatively overmasks. The lower Dice in the crowd test is therefore not a contradiction of the main results; it reflects the harder privacy policy and the fact that \sys prefers removing extra nearby pixels over exposing private people.

\begin{table}[t]
\centering
\caption{Crowd stress test with different public/private compositions. Values are averaged across scenes with three or more people.}
\label{tab:mixed_people_main}
\small
\begin{tabular}{lcc}
\toprule
Composition & Dice $\uparrow$ & Recall $\uparrow$ \\
\midrule
1 public / 2 private & 0.695 & 0.932 \\
2 public / 1 private & 0.684 & 0.799 \\
2 public / 2 private & 0.656 & 0.960 \\
\bottomrule
\end{tabular}
\end{table}

\paragraph{Comparison with segmentation baselines.} We compare to SAM \citep{kirillov2023segment} and a lightweight Tiny U-Net with a ResNet-18 encoder \citep{yakubovskiy2019segmentation}, EdgeSAM~\citep{zhou2023edgesam}, and EdgeTAM~\citep{zhou2025edgetam}. These baselines are useful segmentation references, but they do not address the full privacy policy because they do not know which same-class instance should remain public. Table~\ref{tab:baselines} therefore evaluates private-object segmentation accuracy and latency. \sys is close to SAM in Dice while being much faster, and it outperforms lightweight models (Tiny U-Net, EdgeSAM, and EdgeTAM) in Dice while still remaining lightweight.

\begin{table}[t]
\centering
\caption{Baseline comparison and private-object detection after masking. Latency is for the private-object segmentation/masking stage on synthetic data. PODR is the private-person detection rate after masking; before masking, PODR is 100\% in all listed groups.}
\label{tab:baselines}
\small
\resizebox{\linewidth}{!}{%
\begin{tabular}{lcc}
\toprule
Method / setting & Main metric & Runtime or PODR \\
\midrule
Tiny U-Net (ResNet-18) \citep{yakubovskiy2019segmentation} & Dice 56.0\% & 13 ms \\
SAM \citep{kirillov2023segment} & Dice 85.0\% & 1420 ms \\
EdgeSAM \citep{zhou2023edgesam} & Dice 71\% & 98 ms \\
EdgeTAM \citep{zhou2025edgetam} & Dice 69\% & 81 ms \\
\textbf{\sys} & Dice 80.0\% & 90 ms \\
\midrule
Synthetic scenes after \sys & PODR 6.3\% & 1.0--12.5\% range \\
Real scenes after \sys & PODR 14.3\% & 1.0--25.0\% range \\
\bottomrule
\end{tabular}
}%
\end{table}

\paragraph{Privacy attack by private-person detection.} We run a Faster R-CNN person detector on the output to test whether the private person is still detectable. Before masking, the detector finds the private person in every evaluated scene. After \sys, PODR drops to an average of 6.3\% on synthetic scenes and 14.3\% on real scenes (Table~\ref{tab:baselines}). Table~\ref{tab:podr_main} gives the per-scene breakdown. We use this as a pessimistic privacy test because even detecting the existence of a private person counts as leakage.

\begin{table}[t]
\centering
\caption{Private-object detection rates (PODR) in multi-person scenes before and after \sys. Lower PODR means that the private person is less detectable after source-side masking.}
\label{tab:podr_main}
\small
\resizebox{\linewidth}{!}{%
\begin{tabular}{lcc}
\toprule
Scene & Without \sys & With \sys \\
\midrule
Gallery House (synthetic) & 100\% & 5.5\% \\
Gallery Round (synthetic) & 100\% & 1.0\% \\
Jedi Council (synthetic) & 100\% & 12.5\% \\
Conference room (real) & 100\% & 21.0\% \\
Living room (real) & 100\% & 25.0\% \\
Patio (real) & 100\% & 7.1\% \\
Office (real) & 100\% & 1.0\% \\
Floor corridor (real) & 100\% & 17.2\% \\
\bottomrule
\end{tabular}
}%
\end{table}

\begin{figure}[t]
\centering
\begin{tikzpicture}
    \begin{axis}[
        width=0.92\linewidth,
        height=0.44\linewidth,
        xlabel={Chunk size $N$},
        ylabel={FPS},
        legend style={font=\scriptsize, at={(0.5,1.02)}, anchor=south, legend columns=2},
        ymin=0, ymax=240,
        xtick={1,2,5,10,20},
        grid=both,
        grid style={dashed, gray!30},
        tick label style={font=\scriptsize},
        label style={font=\scriptsize}
    ]
        \addplot[mark=*, color=blue] coordinates {(1,12.9) (2,25.0) (5,57.5) (10,112.7) (20,221.7)};
        \addlegendentry{MobileNet}
        \addplot[mark=square*] coordinates {(1,2.3) (2,4.5) (5,11.1) (10,21.6) (20,41.2)};
        \addlegendentry{ResNet-50}
    \end{axis}
\end{tikzpicture}
\caption{Frame rate as a function of chunk size. Larger chunks increase throughput but reduce detector refresh frequency.}
\label{fig:fps_comparison}
\end{figure}
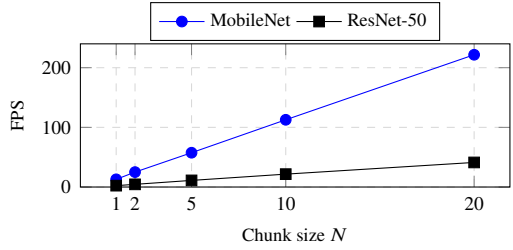

\paragraph{Latency and throughput.} Chunking is the main efficiency lever. Table~\ref{tab:latency_main} reports representative operating points. With MobileNet, masking latency falls from 77.5 ms at $N=1$ to 17.4 ms at $N=5$, while throughput increases from 12.9 FPS to 57.5 FPS. With ResNet-50, throughput reaches 41.2 FPS at $N=20$, but recall decreases as chunks become too long. Figures~\ref{fig:masking_latency}--\ref{fig:fps_comparison} report the full latency and FPS curves, while Table~\ref{tab:backbone} reports the backbone trade-off. In practice, $N$ should be chosen from the expected speed of private objects: smaller chunks favor privacy under fast motion, while larger chunks favor throughput in static or slow scenes.

\begin{table}[t]
\centering
\caption{Representative latency/throughput operating points. Masking latency is measured on the edge device; E2E latency includes the full masked point-cloud pipeline.}
\label{tab:latency_main}
\small
\resizebox{\linewidth}{!}{%
\begin{tabular}{lcccc}
\toprule
Backbone, chunk & Mask ms $\downarrow$ & E2E ms $\downarrow$ & FPS $\uparrow$ & Main trade-off \\
\midrule
MobileNet, $N=1$ & 77.5 & 357.7 & 12.9 & maximum refresh \\
MobileNet, $N=5$ & 17.4 & 297.6 & 57.5 & balanced edge mode \\
ResNet-50, $N=5$ & 90.0 & 370.2 & 11.1 & higher accuracy \\
ResNet-50, $N=20$ & 24.3 & 304.4 & 41.2 & high throughput \\
\bottomrule
\end{tabular}
}%
\end{table}

\paragraph{Backbone and refresh trade-offs.} To preserve the existing latency evidence in the main paper, Table~\ref{tab:backbone} reports the accuracy trade-off between the MobileNet and ResNet-50 backbones on the real dataset, and Figures~\ref{fig:masking_latency}--\ref{fig:fps_comparison} show how chunk size changes masking latency, end-to-end latency, and frame rate.

\begin{table}[t]
\centering
\caption{Backbone accuracy trade-off on the real dataset.}
\label{tab:backbone}
\small
\begin{tabular}{lcc}
\toprule
Backbone & Dice & Recall \\
\midrule
MobileNet & 0.80 & 0.87 \\
ResNet-50 & 0.82 & 0.89 \\
\bottomrule
\end{tabular}
\end{table}

\begin{figure}[t]
\centering
\begin{tikzpicture}
    \begin{axis}[
        width=0.92\linewidth,
        height=0.44\linewidth,
        xlabel={Chunk size $N$},
        ylabel={Latency (ms)},
        legend style={font=\scriptsize, at={(0.5,1.02)}, anchor=south, legend columns=2},
        ymin=0, ymax=800,
        xtick={1,2,5,10,20},
        grid=both,
        grid style={dashed, gray!30},
        tick label style={font=\scriptsize},
        label style={font=\scriptsize}
    ]
        \addplot[mark=diamond*, color=blue] coordinates {(1,77.5) (2,40) (5,17.38) (10,8.87) (20,4.51)};
        \addlegendentry{MobileNet masking}
        \addplot[mark=*, color=orange] coordinates {(1,436.6) (2,221.95) (5,90.02) (10,46.35) (20,24.27)};
        \addlegendentry{ResNet-50 masking}
        \addplot[mark=square*, color=magenta] coordinates {(1,357.67) (2,320.17) (5,297.55) (10,289.04) (20,284.68)};
        \addlegendentry{MobileNet E2E}
        \addplot[mark=triangle*] coordinates {(1,716.77) (2,502.12) (5,370.19) (10,326.52) (20,304.43)};
        \addlegendentry{ResNet-50 E2E}
    \end{axis}
\end{tikzpicture}
\caption{Masking and end-to-end latency for two backbones as a function of chunk size.}
\label{fig:masking_latency}
\end{figure}
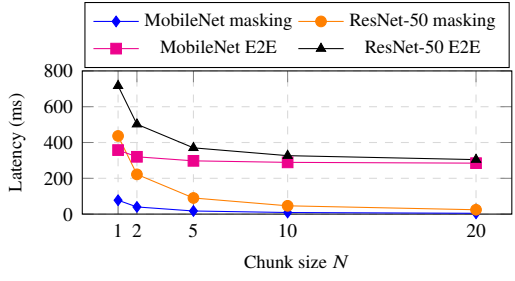

\paragraph{Summary.} The results support three application-level conclusions. First, source-side filtering is feasible for volumetric streaming: \sys removes private RGB and depth content before cloud fusion while preserving public scene structure, as reflected by high recall and synthetic SSIM above 0.98. Second, the task is not equivalent to generic segmentation. The crowd stress test shows that same-class public/private ambiguity is central: a class-level masker would either remove all people or leak private people, while \sys uses reference-view public-instance transfer to keep declared public participants and remove private ones. Third, the runtime results show a practical operating range rather than a single fixed point. Smaller chunks refresh detections more often and favor privacy under motion, while larger chunks improve throughput for static or slow scenes. This lets deployments select the privacy/latency trade-off according to expected scene dynamics without changing the privacy policy.

\paragraph{Application implications.} The evaluation also clarifies where \sys fits in an end-to-end VVS deployment. The camera-side module does not need to solve full scene understanding; it only needs to enforce a local privacy policy before raw RGB-D data crosses the trust boundary. This is important for room-scale capture, where a public participant, a private bystander, and private objects may all appear in the same sensitive class set. A deployment can therefore treat the reference view and working area as a simple policy interface: content inside the public region is retained, while same-class content outside the policy is removed unless it can be matched consistently across views. The resulting sanitized frames remain compatible with the downstream point-cloud pipeline, so the privacy mechanism changes the data admitted to the cloud rather than requiring a new renderer or streaming format.

\section{Robustness and Limitations}
\label{sec:robustness}

\begin{figure}[t]
  \centering
  \includegraphics[width=1\linewidth]{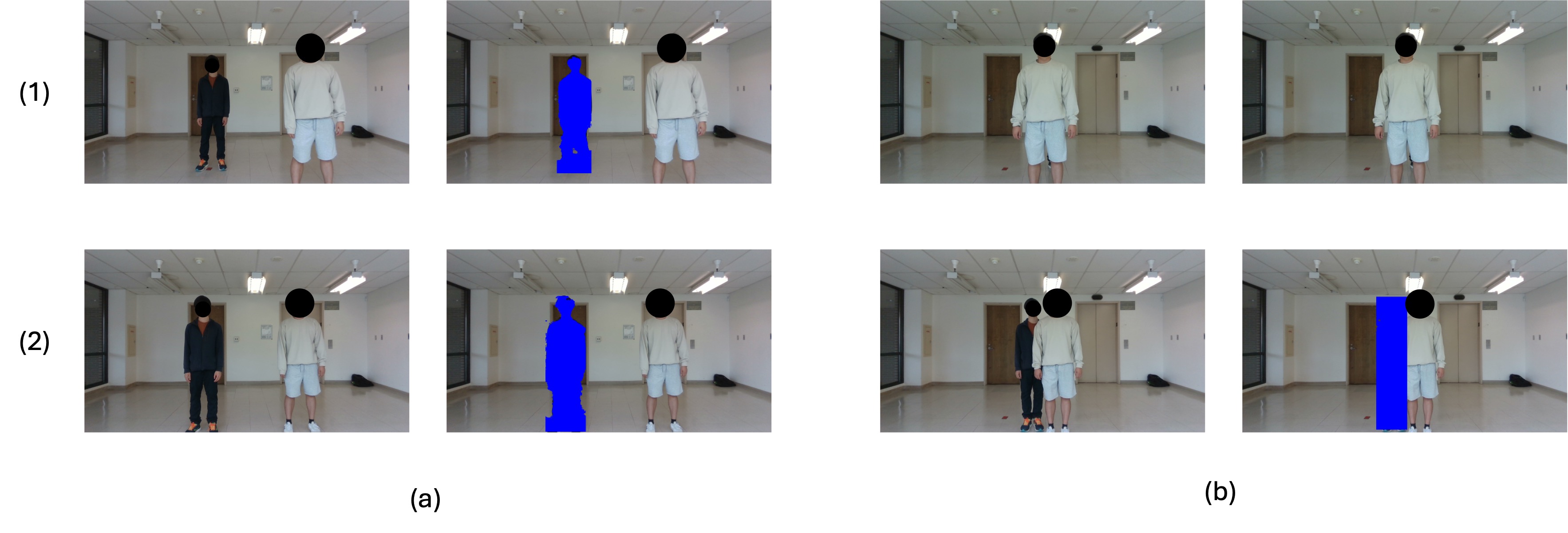}
  \caption{\textbf{Geometry stress cases.} The four cases vary lateral overlap and depth separation between public and private people. \sys uses both the bounding-box constraint and the depth profile, then uses reference-view public/private transfer when same-class objects are close or overlapping.}
  \label{fig:geom_main}
\end{figure}

\paragraph{Geometry.} \sys handles several common public/private layouts: separated objects at different depths, overlapping boxes at different depths, separated boxes at similar depths, and the worst case of overlapping boxes at similar depths. The depth term separates objects when depths differ, and the bounding-box term limits spillage when depths are similar. In the worst case, reference-view public selection and cross-view transfer are essential: the matched public box is excluded from masking and all other sensitive boxes are treated as private. Figure~\ref{fig:geom_main} visualizes these four cases. The key point is that no single cue is reliable in all layouts: depth separates objects when their boxes overlap, boxes localize the mask when depths are similar, and multi-view public-instance transfer resolves the hardest same-class ambiguity.

\paragraph{Motion.} \sys is also capable of handling more complex cases such as motion. Let $\bar{v}$ be the average speed of a private object along the depth axis during one detection cycle. If frames arrive every $\Delta t$ and the detector runs every $N$ frames, the depth displacement is $\delta_z=\bar{v}N\Delta t$. A reused depth mask remains valid while $\delta_z\le \epsilon\tau$, where $\epsilon$ is a tolerance factor and $\tau$ is the depth threshold. Thus, fast motion requires smaller chunks. Empirically, Dice/Recall decrease from 0.808/0.904 at $N=1$ to 0.695/0.735 at $N=20$ with ResNet-50 on synthetic data.

\paragraph{Why existing privacy and segmentation methods are insufficient.}
Existing approaches are not directly applicable to volumetric video streaming. Most prior work in visual privacy focuses on single-camera, 2D, frame-based settings~\cite{wu2021pecam,lu2022preva,yu2018pinto,singh2021disco,padilla2015visual,mireshghallah2020shredder,van2022client,jiang2022primask,kim2017viewmap,ilia2015face,zhu2023campro,shan2020fawkes}. These methods typically operate independently per frame or per view and do not reason about cross-view geometric consistency.

State-of-the-art segmentation models, including large models such as SAM~\cite{kirillov2023segment} and edge-friendly variants such as EdgeSAM~\cite{zhou2023edgesam}, Tiny U-Net~\cite{yakubovskiy2019segmentation}, and EdgeTAM~\cite{zhou2025edgetam}, are similarly \textit{ill-suited} for VVS deployments. \textit{First}, they are not designed for 3D capture setups involving dynamic and overlapping objects observed simultaneously from multiple viewpoints. Applying segmentation independently to each camera feed can lead to inconsistent masking across views, which manifests as visual artifacts or partial privacy leakage after 3D reconstruction. \textit{Second}, dense per-frame segmentation incurs substantial computational overhead, making it difficult to meet real-time constraints on resource-constrained edge devices commonly used in multi-camera VVS systems. \textit{Third}, segmentation accuracy alone does not offer privacy in a volumetric pipeline: the masking needs to know which objects to remove and which ones to keep. This would require additional information about the \textit{location} of objects. 

More broadly, privacy challenges in VVS are fundamentally system-level problems. They involve where masking is applied (edge vs.\ cloud), how masking decisions propagate across views, how masked data interacts with 3D reconstruction, and how privacy protections trade off against latency, throughput, and user experience. These considerations are largely orthogonal to improving segmentation accuracy in isolation and motivate the need for a dedicated, volumetric-aware privacy mechanism.

\paragraph{Limitations.}
\label{sec:limitations}
\sys depends on detector quality, depth quality, and camera calibration. Commodity RGB-D sensors can fail around reflective surfaces, transparent objects, low light, direct sunlight, or long distances. Calibration errors can misproject public centers across views, especially when multiple people are close together. The current real-world evaluation uses sequential static capture epochs rather than fully synchronized multi-camera hardware; this validates the multi-view geometry and masking logic but does not fully measure synchronization errors in a deployed camera array. Finally, \sys removes visual and depth content but does not protect audio, metadata, or private information already visible on public objects. A deployment risk is false confidence: users may assume that all private information has been removed even when detector, depth, or calibration failures leave residual evidence. These limitations are consistent with the threat model and are important directions for deployment hardening.

\paragraph{Deployment implications.}
InViStream is designed for room-scale capture rather than a closed-world segmentation benchmark. The system must decide what data is allowed to leave the source while preserving public scene content; simply removing every detected person would harm utility, while preserving every person would leak private bystanders. It therefore targets the middle case: public same-class instances remain visible, while sensitive instances are removed before fusion.

\paragraph{Interpreting the metrics.}
Dice, recall, PODR, SSIM, point distance, latency, and FPS answer different parts of the deployment question. Dice measures how tightly the private mask matches the ground truth, but recall is the more privacy-facing mask metric because it measures how much of the private object is removed. PODR is a downstream privacy check: even if a mask has reasonable overlap, a remaining silhouette can still be detected as a person after fusion. SSIM and point distance protect against the opposite failure mode, where the system hides the private object by damaging the public scene. Finally, latency and FPS show whether the method can run before transmission rather than only as an offline post-processing step. Taken together, the metrics support the central claim that source-side privacy for VVS is a privacy/utility/runtime trade-off, not a single segmentation score.

\paragraph{Why multi-view reasoning is necessary.}
The hardest cases in the paper are not simply low-resolution masks. They occur when private and public objects share the same semantic class, overlap in one camera, or appear differently across views. In these cases, a single RGB frame may not contain enough information to decide whether a detected person should remain public or be removed. Depth helps separate objects that overlap in image space, but depth alone is not enough when objects are close or have similar range. Calibrated multi-view transfer supplies the missing policy information: public instances selected in the reference view are projected into other views, and unmatched same-class detections are treated conservatively as private. This design is what makes \sys specific to volumetric capture rather than a direct reuse of ordinary 2D redaction.

\paragraph{Choosing an operating point.}
The runtime results should be read as a menu of operating points rather than as a single fixed configuration. For static scenes, meetings, or slow background motion, a larger chunk can reduce detector overhead and increase throughput. For faster movement, a smaller chunk refreshes boxes and depth profiles more often, improving privacy recall at higher cost. The system does not require changing the cloud-side renderer or retraining the detector to move along this trade-off; it changes only how often the local privacy state is refreshed. This is important for an application deployment because the same architecture can be tuned for a high-refresh privacy mode or a high-throughput mode depending on the environment.

\paragraph{Scope of the claim.}
The results support source-side filtering under the stated honest-but-curious cloud model. They do not imply absolute privacy, protection against compromised sensors, or robustness to every physical scene artifact. Instead, the contribution is a practical vision pipeline and evaluation setting for a new class of privacy failures introduced by volumetric streaming. The key lesson is that private appearance, private depth geometry, and cross-view consistency must be handled before cloud fusion; otherwise, information removed in one place can be reconstructed or rediscovered through another view.

\section{Related Work}

VVS systems improve reconstruction, delivery, compression, and mixed-reality quality. Holoportation reconstructs remote 3D scenes in real time \citep{orts2016holoportation}; MetaStream, Vues, M5, Habitus, FarFetchFusion, Theia, ImmerScope, MagicStream, and related systems optimize bandwidth, mobility, pose awareness, user experience, or rendering quality \citep{guan2023metastream,liu2022vues,zhang2022m5,zhang2024habitus,lee2023farfetchfusion,wu2024theia,chen2024immerscope,cheng2024magicstream,liu2024muv2}. These systems are complementary to \sys: they decide how to transmit volumetric content efficiently, while \sys decides what content may safely leave the source.

Privacy-preserving image analysis includes transformations, obfuscation, secure inference, and in-camera filtering \citep{wu2021pecam,lu2022preva,yu2018pinto,singh2021disco,padilla2015visual,mireshghallah2020shredder,van2022client,jiang2022primask,kim2017viewmap,ilia2015face,zhu2023campro,shan2020fawkes}. These methods motivate our source-side approach, but most are designed for 2D frames, single cameras, or class-level privacy. \sys targets the additional constraints of volumetric streaming: RGB-D data, same-class public/private ambiguity, and consistent multi-view fusion.

\section{Conclusion}

\sys is a source-side privacy system for real-time volumetric video streaming. It combines off-the-shelf object detection, depth-aware masking, multi-view public/private synchronization, and private point removal to sanitize RGB-D data before cloud fusion. The evaluation shows that privacy for volumetric media should be treated as a source-side, multi-view computer-vision problem rather than a post-processing step after reconstruction. The main lesson is that object identity, depth geometry, and calibrated view consistency must be used together: each cue alone fails in common room-scale layouts. \sys provides a practical baseline for privacy-preserving VVS and exposes the key trade-offs among privacy recall, public-scene utility, and interactive performance. As immersive telepresence and shared-space sensing move into everyday environments, these trade-offs will be central to making volumetric capture usable without exposing bystanders or private objects by default.

\clearpage
\appendix
\input{appendix.tex}

\clearpage
{\small
\bibliographystyle{ieeenat_fullname}
\bibliography{ref}
}
\end{document}

%% file: authors.tex
\author{%
  Hossein Khalili$^{1}$ \quad
  Philip Do$^{1}$ \quad
  Alexander Vilesov$^{1}$\\[-0.05em]
  Achuta Kadambi$^{1}$ \quad
  Kittipat Apicharttrisorn$^{2}$ \quad
  Nader Sehatbakhsh$^{1}$\\[0.45em]
  \small $^{1}$University of California, Los Angeles \qquad
  $^{2}$Nokia Bell Labs
}

%% file: appendix.tex
\section{Implementation Details}
\label{app:implementation}

\paragraph{Synthetic data generation.}
We create synthetic scenes by loading room meshes and human models as point clouds in Open3D. Virtual cameras are placed at eight equally spaced viewpoints and oriented toward the scene center. For each camera, we save RGB-D images, camera intrinsics, extrinsics, and the segmentation ground truth. Ground truth is obtained by assigning object-specific colors in the source 3D models before rendering. This gives exact private/public labels for every rendered point and pixel.

\paragraph{Real data generation.}
The real dataset uses an Intel RealSense D435. Each location is captured from eight calibrated viewpoints and under four activity scenarios: conversation, walking, sitting, and standing. The captures involved six adult participants who provided informed consent for research use of their RGB-D data and anonymized paper figures. Before recording, participants received written study instructions describing the capture procedure, collected data, expected activities, privacy risks, and their right to stop participating. No audio or unrelated personal metadata was collected. Raw real-person RGB-D data is stored in access-controlled project storage and is not released with this preprint. No crowdsourcing platform was used, and participants did not receive task-based crowdsourcing compensation. The data collection was reviewed under an institutional human-subjects process. Since real scenes do not provide object-level ground truth, we use SAM~\cite{kirillov2023segment} with manual prompts to produce segmentation masks for evaluation. The detector used by \sys is not trained on these masks.

\paragraph{Third-party assets and licenses.}
We use third-party software, pretrained models, and external 3D assets only for research evaluation. Open3D is used for 3D processing and is distributed under the MIT License. SAM is used only to generate evaluation masks for real scenes and is distributed under the Apache License 2.0. COCO-pretrained TorchVision detectors are used as off-the-shelf detection backbones; we do not redistribute COCO images or annotations. We use Volograms/V-SENSE and 8i human-model assets only for non-commercial research evaluation under their accompanying dataset license agreements. Synthetic room assets are credited to their original repositories or creators and will not be redistributed unless permitted by the asset-specific license. The real RGB-D captures are governed by the human-subjects protocol described above. Any released package will include only code, data, or assets whose licenses and consent terms permit release, together with an asset manifest listing source, attribution, and license or usage terms.

\paragraph{Hardware.}
Edge-side measurements use an NVIDIA Jetson Orin Nano with a 6-core ARM CPU, 1024-core Ampere-class GPU, 32 Tensor Cores, and 8 GB RAM. Cloud-side point-cloud processing uses Open3D tensor APIs on a server with two NVIDIA A6000 GPUs connected with NVLink, 96 GB GPU memory, and a 10-core Intel CPU for thread management.

\section{Additional Robustness Trade-offs}
\label{app:robustness}
\begin{figure}[H]
\centering
\begin{tikzpicture}
    \begin{axis}[
        width=0.92\linewidth,
        height=0.48\linewidth,
        xlabel={Chunk size $N$},
        ylabel={Score},
        legend style={font=\scriptsize, at={(0.5,1.02)}, anchor=south, legend columns=2},
        ymin=0.65, ymax=1.0,
        xtick={1,2,5,10,20},
        grid=both,
        grid style={dashed, gray!30},
        tick label style={font=\scriptsize},
        label style={font=\scriptsize}
    ]
        \addplot[mark=o] coordinates {(1,0.8080) (2,0.8070) (5,0.7990) (10,0.7760) (20,0.6952)};
        \addlegendentry{Dice}
        \addplot[mark=square*] coordinates {(1,0.9035) (2,0.9022) (5,0.8909) (10,0.8562) (20,0.7345)};
        \addlegendentry{Recall}
    \end{axis}
\end{tikzpicture}
\caption{Impact of chunk size on Dice and Recall. Larger $N$ reduces latency but degrades accuracy when private objects move between detector refreshes.}
\label{fig:appendix_chunks_accuracy}
\end{figure}
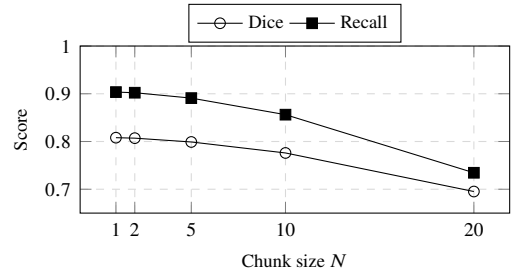

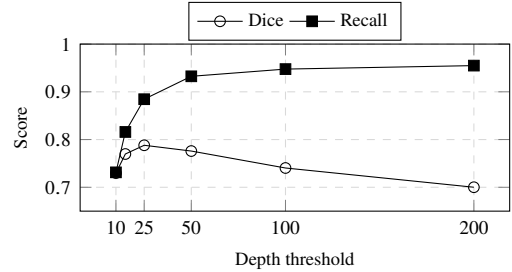
\begin{figure}[H]
\centering
\begin{tikzpicture}
    \begin{axis}[
        width=0.92\linewidth,
        height=0.48\linewidth,
        xlabel={Depth threshold},
        ylabel={Score},
        legend style={font=\scriptsize, at={(0.5,1.02)}, anchor=south, legend columns=2},
        ymin=0.65, ymax=1.0,
        xtick={10,25,50,100,200},
        grid=both,
        grid style={dashed, gray!30},
        tick label style={font=\scriptsize},
        label style={font=\scriptsize}
    ]
        \addplot[mark=o] coordinates {(10,0.7297) (15,0.7697) (25,0.7880) (50,0.7759) (100,0.7404) (200,0.7001)};
        \addlegendentry{Dice}
        \addplot[mark=square*] coordinates {(10,0.7313) (15,0.8159) (25,0.8845) (50,0.9325) (100,0.9476) (200,0.9549)};
        \addlegendentry{Recall}
    \end{axis}
\end{tikzpicture}
\caption{Sensitivity to the depth-threshold parameter. Increasing the threshold improves recall by overmasking but can reduce Dice by removing non-private content.}
\label{fig:appendix_depth_threshold}
\end{figure}

Figures~\ref{fig:appendix_chunks_accuracy} and~\ref{fig:appendix_depth_threshold} report the main robustness trade-offs. Figure~\ref{fig:appendix_chunks_accuracy} shows the effect of chunk size on Dice and Recall, while Figure~\ref{fig:appendix_depth_threshold} shows the effect of the depth-threshold parameter.

\clearpage
\onecolumn
\section{Algorithms}
\label{app:algorithms}

\begin{algorithm}[H]
\caption{Depth-based privacy masking with multi-view synchronization.}
\label{alg:masking}
\small
\begin{algorithmic}[1]
\REQUIRE RGB-D frames $\{I_{v,t},D_{v,t}\}$, reference view $r$, detection interval $N$, threshold $\theta$, depth window size $w$, multiplier $\alpha$, sensitive classes $\mathcal{C}$, transforms $\mathbf{T}_{r\rightarrow v}$
\STATE Initialize masks $M_{v,t}\gets 0$ and stored depth profiles $\mathcal{S}_v\gets\emptyset$.
\FOR{$t=1$ to $T$}
  \IF{$t \bmod N = 1$}
    \STATE Detect candidates $\mathcal{O}_{r,t}=\textsc{Detect}(I_{r,t})$ in the reference view.
    \STATE Mark sensitive boxes inside the public working area as public; mark other same-class boxes as private.
    \STATE Extract each public center $(r_x,r_y,r_z)$ in 3D.
    \FOR{$v=1$ to $V$}
      \STATE Detect candidates $\mathcal{O}_{v,t}=\textsc{Detect}(I_{v,t})$.
      \STATE Transfer each public center to view $v$ using $\mathbf{T}_{r\rightarrow v}$ and mark the nearest same-class box as public.
      \STATE Mark all remaining boxes with labels in $\mathcal{C}$ as private.
      \STATE For every private box, compute $(\mu,\sigma)$ from the center depth window and store $(b,\mu,\alpha\sigma)$ in $\mathcal{S}_v$.
    \ENDFOR
  \ENDIF
  \FOR{$v=1$ to $V$}
    \FOR{each $(b,\mu,\tau)\in\mathcal{S}_v$}
      \STATE $M_{v,t}\gets M_{v,t}\cup \{(u,v')\in b: |D_{v,t}(u,v')-\mu|\le\tau\}$.
    \ENDFOR
  \ENDFOR
\ENDFOR
\RETURN Masks $\{M_{v,t}\}$.
\end{algorithmic}
\end{algorithm}

\begin{algorithm}[H]
\caption{Multi-view private point removal and point-cloud merging.}
\label{alg:merge}
\small
\begin{algorithmic}[1]
\REQUIRE Sanitized RGB-D frames $\{I_{v,t},D_{v,t}\}$, camera extrinsics $\{E_v\}$, private marker color $c_{priv}$
\FOR{$t=1$ to $T$}
  \STATE $\mathcal{P}^{merged}_t\gets\emptyset$.
  \FOR{$v=1$ to $V$}
    \STATE $p\gets\textsc{ConvertToPointCloud}(I_{v,t},D_{v,t})$.
    \FORALL{points $q\in p$}
      \IF{$\textsc{Color}(q)=c_{priv}$}
        \STATE $\textsc{Position}(q)\gets(\infty,\infty,\infty)$.
      \ENDIF
    \ENDFOR
    \STATE Remove non-finite points from $p$.
    \STATE $p\gets\textsc{ApplyTransform}(p,E_v)$.
    \STATE $\mathcal{P}^{merged}_t\gets\mathcal{P}^{merged}_t\cup p$.
  \ENDFOR
\ENDFOR
\RETURN $\{\mathcal{P}^{merged}_t\}_{t=1}^{T}$.
\end{algorithmic}
\end{algorithm}

\clearpage
\twocolumn

%% file: ref.bib
@inproceedings{orts2016holoportation,
  title={Holoportation: Virtual 3d teleportation in real-time},
  author={Orts-Escolano, Sergio and Rhemann, Christoph and Fanello, Sean and Chang, Wayne and Kowdle, Adarsh and Degtyarev, Yury and Kim, David and Davidson, Philip L and Khamis, Sameh and Dou, Mingsong and others},
  booktitle={Proceedings of the 29th annual symposium on user interface software and technology},
  pages={741--754},
  year={2016}
}

@article{zhou2023edgesam,
  title={Edgesam: Prompt-in-the-loop distillation for on-device deployment of sam},
  author={Zhou, Chong and Li, Xiangtai and Loy, Chen Change and Dai, Bo},
  journal={arXiv preprint arXiv:2312.06660},
  year={2023}
}

@inproceedings{zhou2025edgetam,
  title={EdgeTAM: On-Device Track Anything Model},
  author={Zhou, Chong and Zhu, Chenchen and Xiong, Yunyang and Suri, Saksham and Xiao, Fanyi and Wu, Lemeng and Krishnamoorthi, Raghuraman and Dai, Bo and Loy, Chen Change and Chandra, Vikas and others},
  booktitle={Proceedings of the Computer Vision and Pattern Recognition Conference},
  pages={13832--13842},
  year={2025}
}

@inproceedings{wu2024theia,
  title={Theia: Gaze-driven and Perception-aware Volumetric Content Delivery for Mixed Reality Headsets},
  author={Wu, Nan and Liu, Kaiyan and Cheng, Ruizhi and Han, Bo and Zhou, Puqi},
  booktitle={Proceedings of the 22nd Annual International Conference on Mobile Systems, Applications and Services},
  pages={70--84},
  year={2024}
}

@inproceedings{zhang2024habitus,
  title={Habitus: Boosting Mobile Immersive Content Delivery through Full-body Pose Tracking and Multipath Networking},
  author={Zhang, Anlan and Wang, Chendong and Hu, Yuming and Hassan, Ahmad and Zhang, Zejun and Han, Bo and Qian, Feng and Xu, Shichang},
  booktitle={21st USENIX Symposium on Networked Systems Design and Implementation (NSDI 24)},
  pages={1677--1695},
  year={2024}
}

@inproceedings{zhang2022m5,
  title={M5: Facilitating multi-user volumetric content delivery with multi-lobe multicast over mmWave},
  author={Zhang, Ding and Zhou, Puqi and Han, Bo and Pathak, Parth},
  booktitle={Proceedings of the 20th ACM Conference on Embedded Networked Sensor Systems},
  pages={31--46},
  year={2022}
}

@inproceedings{zhang2021efficient,
  title={Efficient volumetric video streaming through super resolution},
  author={Zhang, Anlan and Wang, Chendong and Han, Bo and Qian, Feng},
  booktitle={Proceedings of the 22nd International Workshop on Mobile Computing Systems and Applications},
  pages={106--111},
  year={2021}
}

@inproceedings{guan2023metastream,
  title={Metastream: Live volumetric content capture, creation, delivery, and rendering in real time},
  author={Guan, Yongjie and Hou, Xueyu and Wu, Nan and Han, Bo and Han, Tao},
  booktitle={Proceedings of the 29th Annual International Conference on Mobile Computing and Networking},
  pages={1--15},
  year={2023}
}

@inproceedings{lee2023farfetchfusion,
  title={Farfetchfusion: Towards fully mobile live 3d telepresence platform},
  author={Lee, Kyungjin and Yi, Juheon and Lee, Youngki},
  booktitle={Proceedings of the 29th Annual International Conference on Mobile Computing and Networking},
  pages={1--15},
  year={2023}
}

@inproceedings{liu2022vues,
  title={Vues: Practical mobile volumetric video streaming through multiview transcoding},
  author={Liu, Yu and Han, Bo and Qian, Feng and Narayanan, Arvind and Zhang, Zhi-Li},
  booktitle={Proceedings of the 28th Annual International Conference on Mobile Computing And Networking},
  pages={514--527},
  year={2022}
}

@inproceedings{han2020vivo,
  title={ViVo: Visibility-aware mobile volumetric video streaming},
  author={Han, Bo and Liu, Yu and Qian, Feng},
  booktitle={Proceedings of the 26th annual international conference on mobile computing and networking},
  pages={1--13},
  year={2020}
}

@inproceedings{chen2024immerscope,
  title={ImmerScope: Multi-view Video Aggregation at Edge towards Immersive Content Services},
  author={Chen, Bo and Guo, Hongpeng and Wu, Mingyuan and Yang, Zhe and Yan, Zhisheng and Nahrstedt, Klara},
  booktitle={Proceedings of the 22nd ACM Conference on Embedded Networked Sensor Systems},
  pages={82--96},
  year={2024}
}

@inproceedings{cheng2024magicstream,
  title={MagicStream: Bandwidth-conserving Immersive Telepresence via Semantic Communication},
  author={Cheng, Ruizhi and Wu, Nan and Le, Vu and Chai, Eugene and Varvello, Matteo and Han, Bo},
  booktitle={Proceedings of the 22nd ACM Conference on Embedded Networked Sensor Systems},
  pages={365--379},
  year={2024}
}

@inproceedings{liu2024muv2,
  title={MuV2: Scaling up Multi-user Mobile Volumetric Video Streaming via Content Hybridization and Sharing},
  author={Liu, Yu and Zhou, Puqi and Zhang, Zejun and Zhang, Anlan and Han, Bo and Li, Zhenhua and Qian, Feng},
  booktitle={Proceedings of the 30th Annual International Conference on Mobile Computing and Networking},
  pages={327--341},
  year={2024}
}

@inproceedings{wu2021pecam,
  title={PECAM: privacy-enhanced video streaming and analytics via securely-reversible transformation},
  author={Wu, Hao and Tian, Xuejin and Li, Minghao and Liu, Yunxin and Ananthanarayanan, Ganesh and Xu, Fengyuan and Zhong, Sheng},
  booktitle={Proceedings of the 27th Annual International Conference on Mobile Computing and Networking},
  pages={229--241},
  year={2021}
}

@inproceedings{lu2022preva,
  title={Preva: Protecting Inference Privacy through Policy-based Video-frame Transformation},
  author={Lu, Rui and Shi, Siping and Wang, Dan and Hu, Chuang and Zhang, Bihai},
  booktitle={2022 IEEE/ACM 7th Symposium on Edge Computing (SEC)},
  pages={175--188},
  year={2022},
  organization={IEEE}
}

@inproceedings{yu2018pinto,
  title={Pinto: enabling video privacy for commodity iot cameras},
  author={Yu, Hyunwoo and Lim, Jaemin and Kim, Kiyeon and Lee, Suk-Bok},
  booktitle={Proceedings of the 2018 ACM SIGSAC Conference on Computer and Communications Security},
  pages={1089--1101},
  year={2018}
}

@inproceedings{singh2021disco,
  title={DISCO: Dynamic and Invariant Sensitive Channel Obfuscation for deep neural networks},
  author={Singh, Abhishek and Chopra, Ayush and Garza, Ethan and Zhang, Emily and Vepakomma, Praneeth and Sharma, Vivek and Raskar, Ramesh},
  booktitle={Proceedings of the IEEE/CVF Conference on Computer Vision and Pattern Recognition},
  pages={12125--12135},
  year={2021}
}

@article{padilla2015visual,
  title={Visual privacy protection methods: A survey},
  author={Padilla-L{\'o}pez, Jos{\'e} Ram{\'o}n and Chaaraoui, Alexandros Andre and Fl{\'o}rez-Revuelta, Francisco},
  journal={Expert Systems with Applications},
  volume={42},
  number={9},
  pages={4177--4195},
  year={2015},
  publisher={Elsevier}
}

@inproceedings{mireshghallah2020shredder,
  title={Shredder: Learning noise distributions to protect inference privacy},
  author={Mireshghallah, Fatemehsadat and Taram, Mohammadkazem and Ramrakhyani, Prakash and Jalali, Ali and Tullsen, Dean and Esmaeilzadeh, Hadi},
  booktitle={Proceedings of the Twenty-Fifth International Conference on Architectural Support for Programming Languages and Operating Systems},
  pages={3--18},
  year={2020}
}

@inproceedings{van2022client,
  title={Client-optimized algorithms and acceleration for encrypted compute offloading},
  author={van der Hagen, McKenzie and Lucia, Brandon},
  booktitle={Proceedings of the 27th ACM International Conference on Architectural Support for Programming Languages and Operating Systems},
  pages={683--696},
  year={2022}
}

@inproceedings{jiang2022primask,
  title={PriMask: Cascadable and Collusion-Resilient Data Masking for Mobile Cloud Inference},
  author={Jiang, Linshan and Song, Qun and Tan, Rui and Li, Mo},
  booktitle={Proceedings of the 20th ACM Conference on Embedded Networked Sensor Systems},
  pages={164--178},
  year={2022}
}

@inproceedings{kim2017viewmap,
  title={$\{$ViewMap$\}$: Sharing Private $\{$In-Vehicle$\}$ Dashcam Videos},
  author={Kim, Minho and Lim, Jaemin and Yu, Hyunwoo and Kim, Kiyeon and Kim, Younghoon and Lee, Suk-Bok},
  booktitle={14th USENIX Symposium on Networked Systems Design and Implementation (NSDI 17)},
  pages={163--176},
  year={2017}
}

@inproceedings{ilia2015face,
  title={Face/off: Preventing privacy leakage from photos in social networks},
  author={Ilia, Panagiotis and Polakis, Iasonas and Athanasopoulos, Elias and Maggi, Federico and Ioannidis, Sotiris},
  booktitle={Proceedings of the 22nd ACM SIGSAC Conference on computer and communications security},
  pages={781--792},
  year={2015}
}

@inproceedings{zhu2023campro,
  title={Campro: Camera-based anti-facial recognition},
  author={Zhu, Wenjun and Sun, Yuan and Liu, Jiani and Cheng, Yushi and Ji, Xiaoyu and Xu, Wenyuan},
  booktitle={NDSS},
  year={2023}
}

@inproceedings{shan2020fawkes,
  title={Fawkes: Protecting privacy against unauthorized deep learning models},
  author={Shan, Shawn and Wenger, Emily and Zhang, Jiayun and Li, Huiying and Zheng, Haitao and Zhao, Ben Y},
  booktitle={29th USENIX security symposium (USENIX Security 20)},
  pages={1589--1604},
  year={2020}
}

@inproceedings{Redmon2017YOLO9000,
  title={YOLO9000: Better, Faster, Stronger},
  author={Redmon, Joseph and Farhadi, Ali},
  booktitle={Proceedings of the IEEE Conference on Computer Vision and Pattern Recognition},
  pages={7263--7271},
  year={2017}
}

@inproceedings{ren2015faster,
  title={Faster R-CNN: Towards Real-Time Object Detection with Region Proposal Networks},
  author={Ren, Shaoqing and He, Kaiming and Girshick, Ross and Sun, Jian},
  booktitle={Advances in Neural Information Processing Systems (NIPS)},
  pages={91--99},
  year={2015}
}

@inproceedings{lin2014microsoft,
  title={Microsoft {COCO}: Common Objects in Context},
  author={Lin, Tsung-Yi and Maire, Michael and Belongie, Serge and Hays, James and Perona, Pietro and Ramanan, Deva and Doll{\'a}r, Piotr and Zitnick, C Lawrence},
  booktitle={European conference on computer vision (ECCV)},
  pages={740--755},
  year={2014},
  organization={Springer}
}

@inproceedings{he2016deep,
  title={Deep residual learning for image recognition},
  author={He, Kaiming and Zhang, Xiangyu and Ren, Shaoqing and Sun, Jian},
  booktitle={Proceedings of the IEEE conference on computer vision and pattern recognition},
  pages={770--778},
  year={2016}
}

@inproceedings{lin2017feature,
  title={Feature Pyramid Networks for Object Detection},
  author={Lin, Tsung-Yi and Doll{\'a}r, Piotr and Girshick, Ross and He, Kaiming and Hariharan, Bharath and Belongie, Serge},
  booktitle={Proceedings of the IEEE conference on computer vision and pattern recognition (CVPR)},
  pages={2117--2125},
  year={2017}
}

@inproceedings{howard2019searching,
  title={Searching for MobileNetV3},
  author={Howard, Andrew and Sandler, Mark and Chu, Grace and Chen, Liang-Chieh and Chen, Bo and Tan, Mingxing and Wang, Weijun and Zhu, Yukun and Pang, Ruoming and Vasudevan, Vijay and Le, Quoc V and Adam, Hartwig},
  booktitle={Proceedings of the IEEE/CVF International Conference on Computer Vision (ICCV)},
  pages={1314--1324},
  year={2019}
}

@inproceedings{zerman2019subjective, 
  title     = {Subjective and Objective Quality Assessment for Volumetric Video Compression}, 
  author    = {Zerman, Emin and Gao, Pan and Ozcinar, Cagri and Smolic, Aljosa}, 
  year      = {2019}, 
  booktitle = {{IS\&T} Electronic Imaging, Image Quality and System Performance {XVI}} 
}

@techreport{krivokuca2018voxelized,
  author       = {M. Krivokuca and P. A. Chou and P. Savill},
  title        = {8i Voxelized Surface Light Field (8iVSLF) Dataset},
  type         = {Input Document},
  number       = {m42914},
  institution  = {ISO/IEC JTC1/SC29 WG11 (MPEG)},
  address      = {Ljubljana, Slovenia},
  month        = jul,
  year         = {2018}
}

@misc{sketch,
  author       = {{ElinHohler}},
  title        = {{Sketchfab} Profile and {3D} Scene Assets Used in Synthetic Evaluation},
  howpublished = {\url{https://sketchfab.com/ElinHohler}},
  year         = {2026},
  note         = {Accessed June 2026; asset-specific licenses checked on Sketchfab}
}

@article{open3d,
   author  = {Qian-Yi Zhou and Jaesik Park and Vladlen Koltun},
   title   = {{Open3D}: {A} Modern Library for {3D} Data Processing},
   journal = {arXiv:1801.09847},
   year    = {2018},
}

@inproceedings{kirillov2023segment,
  title={Segment anything},
  author={Kirillov, Alexander and Mintun, Eric and Ravi, Nikhila and Mao, Hanzi and Rolland, Chloe and Gustafson, Laura and Xiao, Tete and Whitehead, Spencer and Berg, Alexander C and Lo, Wan-Yen and others},
  booktitle={Proceedings of the IEEE/CVF International Conference on Computer Vision},
  pages={4015--4026},
  year={2023}
}

@misc{yakubovskiy2019segmentation,
  title        = {Segmentation Models},
  author       = {Pavel Yakubovskiy},
  year         = {2019},
  howpublished = {\url{https://github.com/qubvel/segmentation_models.pytorch}},
}
